\pdfoutput=1
\documentclass[11pt]{article}

\usepackage[]{acl}

\usepackage{times}
\usepackage{latexsym}
\usepackage{graphicx}
\usepackage{amsfonts}
\usepackage{booktabs}
\usepackage{multirow, multicol}
\usepackage{tabularx}
\usepackage{float}

\usepackage[T1]{fontenc}
\usepackage[utf8]{inputenc}

\usepackage{microtype}

\title{Cross-lingual Transfer Learning for Javanese Dependency Parsing}

\author{Fadli Aulawi Al Ghiffari, Ika Alfina, \and Kurniawati Azizah \\
  Faculty of Computer Science, Universitas Indonesia, Depok, Indonesia \\
  \texttt{fadli.aulawi@ui.ac.id} \\ 
    \texttt{\{ika.alfina, kurniawati.azizah\}@cs.ui.ac.id} \\
 }   
\begin{document}
\maketitle
\begin{abstract}

While structure learning achieves remarkable performance in high-resource languages, the situation differs for under-represented languages due to the scarcity of annotated data. This study focuses on assessing the efficacy of transfer learning in enhancing dependency parsing for Javanese—a language spoken by 80 million individuals but characterized by limited representation in natural language processing. We utilized the Universal Dependencies dataset consisting of dependency treebanks from more than 100 languages, including Javanese. We propose two learning strategies to train the model: transfer learning (TL) and hierarchical transfer learning (HTL). While TL only uses a source language to pre-train the model, the HTL method uses a source language and an intermediate language in the learning process. The results show that our best model uses the HTL method, which improves performance with an increase of 10 \% for both UAS and LAS evaluations compared to the baseline model. 
\end{abstract}

\section{Introduction}
\label{s:intro}

Despite over 80 million native speakers of Javanese \cite{Simons2023}, this language is underrepresented in NLP due to a scarcity of annotated resources. Limited works in Javanese have focused on stemmer \cite{Soyusiawaty2020}, POS tagger \cite{Askhabi2020}, sentiment analysis \cite{Tho2021}, and machine translation \cite{Lesatari2021}. However, few have explored language structure prediction, such as dependency parsing. Dependency parsing is a process that makes a structural representation of a sentence \cite{Kubler2009} that produces a structure in the form of a dependency tree represented in a graph consisting of several connected links between words in a sentence. 

Recent work, \citet{alfina2023} created a public gold standard dataset for Javanese with 1000 sentences, published as part of the Universal Dependencies dataset \cite{Zeman2023}. This dataset covers annotation for tokenization, POS tagging, morphological features tagging, and dependency parsing tasks. The most recent parser performance \cite{alfina2023} using this dataset is not satisfactory, with only 77.08\% on Unlabeled Attachment Score (UAS) and 71.21\% on Labeled Attachment Score (LAS). The lack of training data is a typical low-resource problem considered one of the biggest NLP research problems \cite{Ruder2019}.

Transfer learning (TL) involves leveraging a model's knowledge from a high-resource source domain to improve performance on various NLP tasks, particularly in low-resource domains \cite{Weiss2016}, by transferring learned information to target tasks. Inspired by \citet{Maulana2022} that utilizes cross-lingual transfer learning to develop an Indonesian dependency parser, we want to try to replicate its outcome in Javanese with a limited available dataset. Moreover, we also implement hierarchical transfer learning (HTL) with two stages of transfer learning that offer increased flexibility over TL by enabling knowledge transfer between languages with a significant gap \cite{Luo2019}, as demonstrated in diverse applications, including Javanese text-to-speech \cite{Azizah2020} and biomedical named entity recognition models \cite{Chai2022}.

We build the dependency parser model for Javanese by adopting model \cite{Ahmad2019} that uses a self-attention encoder and a graph-based decoder. We utilize the Universal Dependency dataset v1.12 \cite{Zeman2023} that provides dependency treebanks for more than 100 languages, including Javanese. Both TL and HTL use a selection of source languages determined by LangRank \cite{Lin2020}. Specifically, HTL employs Indonesian as an intermediary language, developing from our referenced research \cite{Maulana2022}. The empirical results show that transfer learning improves accuracy with a margin of 10\% compared to the baseline. We also report the word embedding comparison that fastText performs better than the Javanese BERT, Javanese RoBERTa, and multilingual BERT. In summary, the main contributions of this paper are as follows:

\begin{enumerate}
    \item Provide the first study of Javanese dependency parsing using TL and HTL strategy. We report that the HTL method can significantly improve performance compared to the training from scratch method.
    \item Report the investigation of which source language and word embedding performs best for TL and HTL strategy.
\end{enumerate}

\section{Related Works}
\label{s:related}

\subsection{Dependency Parser}

The dependency parser model can be developed using two methods, the transition-based and graph-based methods \cite{Das2020}. The transition-based method works by processing the word order one by one in a given sentence \cite{Jurafsky2020}. Meanwhile, the graph-based method gives a score to each edge of the word relation \cite{Jurafsky2020}, then looks for the best tree formed from the edges with the best scores.

Apart from these two methods, there is an approach in which the parser is built using an encoder-decoder architecture. It was first developed using a BiLSTM encoder and a deep biaffine decoder \cite{Dozat2017}. Encoder variations began to develop using Transformers or self-attention encoders \cite{Vaswani2017}, then subsequent studies modified it using relative positional embedding \cite{Shaw2018}. The first Javanese dependency parser \cite{alfina2023} uses UDPipe \cite{Straka2018}, which also utilizes the biaffine attention mentioned before. 

In the context of transfer learning, it was found that the best combination is a self-attention encoder and a graph-based decoder \cite{Ahmad2019}, which will be used in this research. This combination has been better than other encoder-decoder combinations in cross-lingual transfer learning. 

\subsection{Transfer Learning}

Transfer learning involves leveraging a pre-trained model's knowledge to enhance the performance of other models \cite{Sarkar2022}, addressing resource limitations in low-resource domains. Besides that, hierarchical transfer learning offers a transfer learning method in which a new layer is added before the model is transferred to the low-resource language \cite{Luo2019}. Recent work has shown that transferring multiple times could minimize the dissimilarity between the high-resource and the low-resource domain languages \cite{Azizah2020}.

Transfer learning strategy offers direct capability, which means a model is trained on a source task and then applied without any labeled data from the target task. Specifically on the parsing task, previous research already done by \citet{Kurniawan2021} and \citet{Ahmad2019} for developing an unsupervised parsing model in several languages using only English as its source language. That approach can be improved by adding fine-tuning with the available small dataset from low-resource language. Recent work \cite{Maulana2022} shows the fine-tuning approach is better than the zero-shot one for building a parsing model in another low-resource language, Indonesian. 

\section{Method}
\label{s:method}

This section concerns the model's architecture with the addition of the transfer learning method, the dataset and word embedding used to train the model, and the evaluation method of how the model is evaluated.

% \begin{figure*}[t!]
%     \centering
% 	\includegraphics[width=0.8\linewidth]{archTra.png}
% 	\caption{A simple illustration of the model architecture used}
% 	\label{fig:arch}
% \end{figure*}

\subsection{Model Architecture}
This work uses an encoder-decoder architecture of \citet{Ahmad2019}. No parameter modifications were made to maintain the success of the previous work. Because training and fine-tuning the model involves resources from several different languages, only language-independent labels are used where the subtype of the label is not involved. 

\subsubsection{Encoder} 

We convert the words and POS tags from the sentence into their embedding form. The self-attention encoder \cite{Vaswani2017} in this study received an embedding matrix, which concatenates the word and POS embedding matrices. The encoder produces two matrices, $M$ and $N$. $M$ matrix represents the probability of a word in column $j$ having the head of a word in row $i$. In comparison, the $N$ matrix represents the probability of a word in column $j$ having a label in row $i$.

\subsubsection{Decoder}

The decoder receives the two matrices and processes them in two following processes. First, $M$ is processed with the maximum spanning tree algorithm in the following way:

Let $G = (V, E)$ be a graph constructed using directed weighted graph $M$. In this case, a vertex is a word representation, and an edge represents the dependency score of the two words. Let $w : E \rightarrow \mathbb{R}$ be a function that assigns a weight to each edge in $E$. Then, the maximum spanning tree problem seeks to find a spanning tree $T = (V, E_T)$ of $G$ such that:

\begin{equation}
    T = \arg \max_{T'} \sum_{e \in E_{T'}} w(e)
\end{equation}

\noindent subject to the constraint that $T$ is a tree. Then, a list of head $H$ is generated from all the destination nodes in $E_T$. It can be denoted as:

\begin{equation}
    H = \{d_i \mid \exists (s_i, d_i) \in E_T\}, \enspace i=1, \dots, n
\end{equation}

Meanwhile, $N$ is processed to generate $L$, containing the list of labels with the highest probability for each word. Finally, the $H$ and $L$ arrays are used to build the final resulting tree from this model.

\subsubsection{Word Embedding}

This research used two types of word embedding approaches: the static type in the form of fastText and the contextual type in the form of BERT. The two types were selected to compare which type was most suitable for the Javanese parser model.

We chose fastText because of the similarity with that used in the previous research \cite{Maulana2022}. We also used BERT with two scenarios: using a different word embedding for each language (BERT and RoBERTa) and only one word embedding for all languages (multilingual BERT). The BERT and RoBERTa scenario uses all the languages involved except Croatian due to the unavailable resources. 

\begin{figure}[t!]
    \centering
	\includegraphics[width=.7\linewidth]{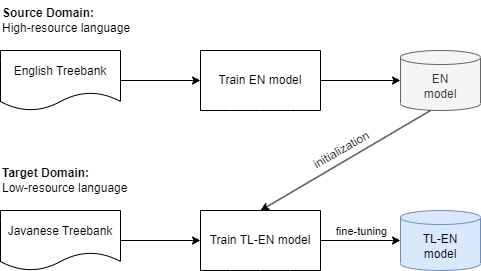}
	\caption{Illustration of standard transfer learning method}
	\label{fig:archTL}
\end{figure}

\begin{figure}[t!]
    \centering
	\includegraphics[width=.7\linewidth]{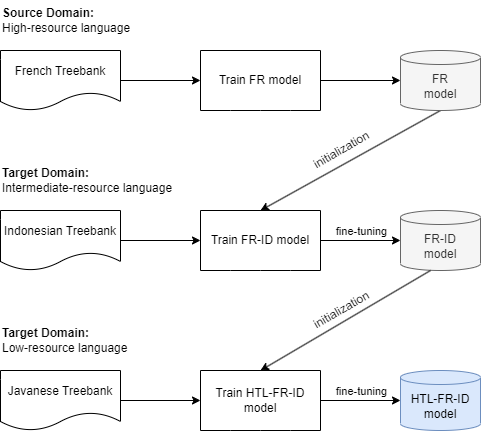}
	\caption{Illustration of hierarchical transfer learning method}
	\label{fig:archHTL}
\end{figure}

\subsection{Training Method}
\label{ss:training}

We perform two training methods: transfer learning and hierarchical transfer learning. Each method generates several models based on the number of source languages used. All models are fine-tuned with the Javanese treebank.

Standard transfer learning only uses one transfer stage from high-resource to low-resource language, as shown in Figure \ref{fig:archTL}. Meanwhile, Figure \ref{fig:archHTL} illustrates a hierarchical transfer learning scenario, where transferring stages are performed twice in hierarchical transfer learning. The first stage is done from a high-source language to an intermediate-resource language, and the second stage is done from an intermediate-resource language to a low-resource language. 

\subsection{Choosing Source Languages}
\label{ss:choosing-source}
Some languages are selected as source languages using the help of LangRank \cite{Lin2020} and references from previous studies. This tool considers combining two main feature groups in each language pair: corpus statistics and typological information.

\subsection{Dataset}

\begin{table}
\caption{The statistics of the Javanese treebank}
  \small
  \begin{tabular}{lr}
    \toprule
    Description & Statistic \\
    \midrule
    Sentence count & 1000 \\ 
    Word count & 14344 \\ 
    Unique word count & 3793 \\
    Average sentence length (in words) & 14.32 \\ 
    Universal Part-of-Speech (UPOS) tag count & 17 \\ 
    Universal dependency relation count & 32 \\
    Language-specific dependency relation count & 14 \\ 
    Total dependency relation count & 46 \\
    \bottomrule
  \end{tabular}
  
  \label{tab:tab-dataset-stat}
\end{table}

\subsubsection{The Javanese dataset}
For the Javanese dataset, we use the only Javanese treebank available in the UD dataset v2.12, the UD\_Javanese-CSUI \cite{alfina2023}. Table~\ref{tab:tab-dataset-stat} shows the statistics of this dataset. The set available for UD\_Javanese-CSUI is only a test set because the data size is still relatively small. We do our split process by following the distribution rule of the data into train, dev, and test sets by 80\%, 10\%, and 10\% percentages. 

\subsubsection{The source language dataset}
\label{sss:sourceLangData}

Langrank recommends the top 3 languages in the following order: Indonesian, Croatian, and Korean. We also use English, one of the important languages in NLP research. These four languages are used in the standard transfer learning scenario. 

For the hierarchical transfer learning scenario using Indonesian as the intermediary language, we choose English, French, and Italy as the source languages suggested by \citet{Maulana2022}. In total, we use six languages as the source languages.

For each source language, we only use one treebank. If a language has more than one treebank in the UD dataset v2.12, we choose the treebank with the biggest size, as shown in Table~\ref{tab:treebankList}.

\begin{table}
\caption{List of treebanks chosen for source languages, with their corresponding size in the number of sentences and words}
  \scriptsize
  \begin{tabular}{lrr}
    \toprule
    Treebank & Sentences & Words \\
    \midrule
    UD\_Croatian-SET \cite{Agic2015} & 9010 & 199409 \\ 
    UD\_English-GUM \cite{Zeldes2017} & 9124 & 164396 \\ 
    UD\_French-GSD \cite{McDonald2013} & 16341 & 400232 \\ 
    UD\_Indonesian-GSD \cite{McDonald2013} & 5598 & 122021 \\ 
    UD\_Italian-ISDT \cite{Bosco2022} & 14167 & 298343 \\ 
    UD\_Korean-GSD \cite{Chun2019} & 6339 & 80322 \\ 
    \bottomrule
  \end{tabular}
  
  \label{tab:treebankList}
\end{table}

\subsection{Experiments Setting}

\subsubsection{Scenarios}
As explained in Section~\ref{ss:training}, we conducted three main scenarios:
\begin{enumerate}
    \item Training from scratch (FS) or baseline scenario, in which the models are trained only using the target language, Javanese.
    \item Standard transfer learning (TL). We construct four distinct models utilizing treebanks from each source language. Then, each model is fine-tuned using the Javanese treebank.
    \item Hierarchical transfer learning (HTL). First, we train three different models using treebank from each source language. After that, the models were fine-tuned with the Indonesian treebank before being fine-tuned again with the Javanese treebank. 
\end{enumerate}

We also compared the performance of the four types of word embeddings for Javanese: fastText \cite{Grave2019}, Javanese BERT \cite{Wongso2021}, Javanese RoBERTa \cite{Wongso2021}, and multilingual BERT \cite{Devlin2019}. 

\subsubsection{Environment}
Implementation is done in Python environments. The training process is supported by the NVIDIA-DGX server with GPU NVIDIA A100 10GB, RAM of 64GB, and storage of 1 TB.

\subsection{Evaluation}

All models are evaluated using the unlabeled attachment score (UAS) and labeled attachment score (LAS) metrics, which are the most frequently used for evaluating the dependency parsing model \cite{Nivre2017}. The margin of error (MOE) with a 95\% confidence level is also used to estimate the range of values within which the true population value is likely to fall. 

\section{Result and Analysis}
\label{s:result}

The evaluation results for all scenarios are shown in Table \ref{tab:resAll}. Scores in bold are marked as the best model in a particular word embedding type metric. 

\begin{table}[ht!]
\caption{Evaluation results of all scenarios}
  \centering
  \scriptsize
    \begin{tabular}{ cccc } 
        \toprule
        Word Embedding & Model & UAS & LAS \\
        \midrule
        \multirow{8}{*}{fastText} & FS & 75.87 ± 2.21 & 68.97 ± 2.39 \\
        \cmidrule{2-4}
         & TL-ID & 84.80 ± 1.85 & 78.10 ± 2.14 \\ 
         & TL-HR & 83.40 ± 1.92 & 76.57 ± 2.19 \\
         & TL-KO & 80.68 ± 2.04 & 74.13 ± 2.26 \\
         & TL-EN & 83.47 ± 1.92 & 77.27 ± 2.16 \\
        \cmidrule{2-4}
         & HTL-EN-ID & 84.94 ± 1.85 & \textbf{79.22 ± 2.10} \\ 
         & HTL-FR-ID & 84.87 ± 1.85 & 77.55 ± 2.15 \\ 
         & HTL-IT-ID & \textbf{85.84 ± 1.80} & 78.87 ± 2.11 \\
        \midrule
        \multirow{8}{*}{jv-BERT} & FS & 74.69 ± 2.25 & 67.29 ± 2.42 \\
        \cmidrule{2-4}
         & TL-ID & 79.08 ± 2.10 & 72.32 ± 2.31 \\ 
         & TL-HR & - & - \\
         & TL-KO & 77.06 ± 2.17 & 70.29 ± 2.36 \\
         & TL-EN & 81.73 ± 2.00 & 75.52 ± 2.22 \\
        \cmidrule{2-4}
         & HTL-EN-ID & \textbf{83.47 ± 1.92} & \textbf{76.64 ± 2.19} \\ 
         & HTL-FR-ID & 81.80 ± 1.99 & 75.38 ± 2.22 \\ 
         & HTL-IT-ID & 81.03 ± 2.02 & 73.99 ± 2.27 \\
        \midrule
        \multirow{8}{*}{jv-RoBERTa} & FS & 69.80 ± 2.37 & 62.97 ± 2.49 \\
        \cmidrule{2-4}
         & TL-ID & 78.45 ± 2.12 & 72.11 ± 2.32 \\ 
         & TL-HR & - & - \\
         & TL-KO & 82.22 ± 1.97 & 76.22 ± 2.20 \\
         & TL-EN & 77.13 ± 2.17 & 70.92 ± 2.35 \\
        \cmidrule{2-4}
         & HTL-EN-ID & 77.41 ± 2.16 & 70.85 ± 2.35 \\ 
         & HTL-FR-ID & 83.05 ± 1.94 & 77.20 ± 2.17 \\ 
         & HTL-IT-ID & \textbf{83.33 ± 1.92} & \textbf{77.20 ± 2.17} \\
        \midrule
        \multirow{8}{*}{multi-BERT} & FS & 75.80 ± 2.21 & 69.04 ± 2.39 \\
        \cmidrule{2-4}
         & TL-ID & 82.01 ± 1.98 & 76.01 ± 2.21 \\ 
         & TL-HR & 83.75 ± 1.90 & 77.68 ± 2.15 \\
         & TL-KO & 79.78 ± 2.07 & 73.29 ± 2.28 \\
         & TL-EN & 80.89 ± 2.03 & 74.13 ± 2.26 \\
        \cmidrule{2-4}
         & HTL-EN-ID & 82.98 ± 1.94 & 76.71 ± 2.18 \\ 
         & HTL-FR-ID & 83.19 ± 1.93 & 77.75 ± 2.15 \\ 
         & HTL-IT-ID & \textbf{84.45 ± 1.87} & \textbf{78.52 ± 2.12} \\
         \bottomrule
    \end{tabular}
  \label{tab:resAll}
\end{table}

\begin{figure}[t!]
    \centering
	\includegraphics[width=.7\linewidth]{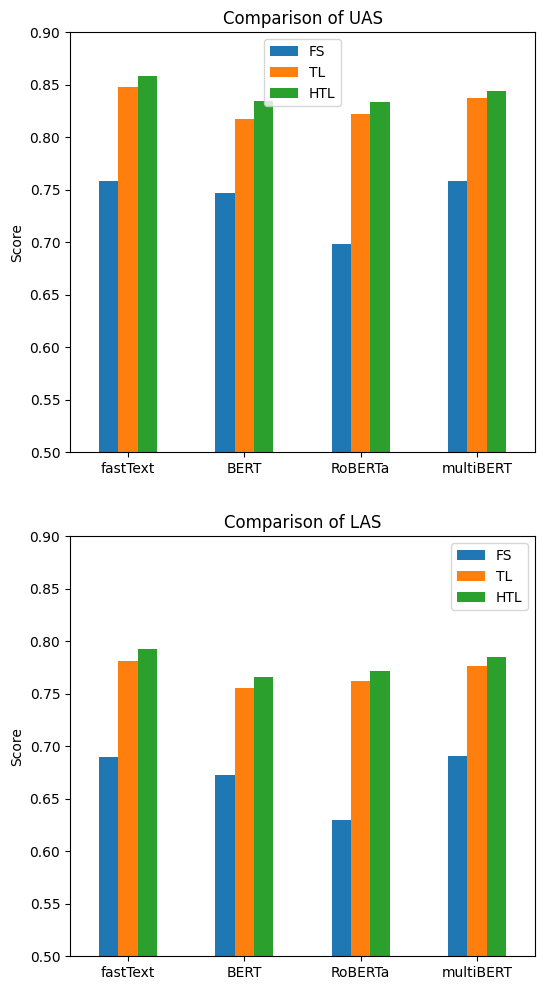}
	\caption{Comparison of the best model evaluation for each scenario}
	\label{fig:compScen}
\end{figure}

\subsection{Models Comparison: From Scratch (FS) Model, Transfer Learning (TL) Model, and Hierarchical Transfer Learning (HTL) Model}

Table \ref{tab:resAll} shows that the transfer learning model performs better than the baseline model in all word embeddings. The performance increase is quite significant, up to 13\% on UAS and 14\% on LAS. This verifies previous studies which explain the advantages of using transfer learning \cite{Sarkar2022}. The lack of resources in Javanese also indicates that transfer learning is suitable for use.

Figure \ref{fig:compScen} also shows that the hierarchical transfer learning method consistently outperforms the transfer learning method even though it is not too significant. Specifically, the comparison focused on the TL-ID and HTL models, as all models from the HTL scenario use the TL-ID model as its second base for the transferring method. The difference between these two scenarios shows that adding suitable high-resource language for the initial source model can give a better performance. 

\subsection{Source Languages Comparison}

Table \ref{tab:resAll} shows that two of the top three recommendations from LangRank have good results. The conclusion is that LangRank can help predict the source language in the Javanese dependency parser. However, it does not rule out the possibility that other languages also have good results. For TL, it cannot be concluded which source language achieves the best performance since different word embedding used by the model gives different results. For HTL using Indonesian as the intermediate language, Italy performs best, followed by English as the source language.

\begin{figure}[t!]
    \centering
	\includegraphics[width=.7\linewidth]{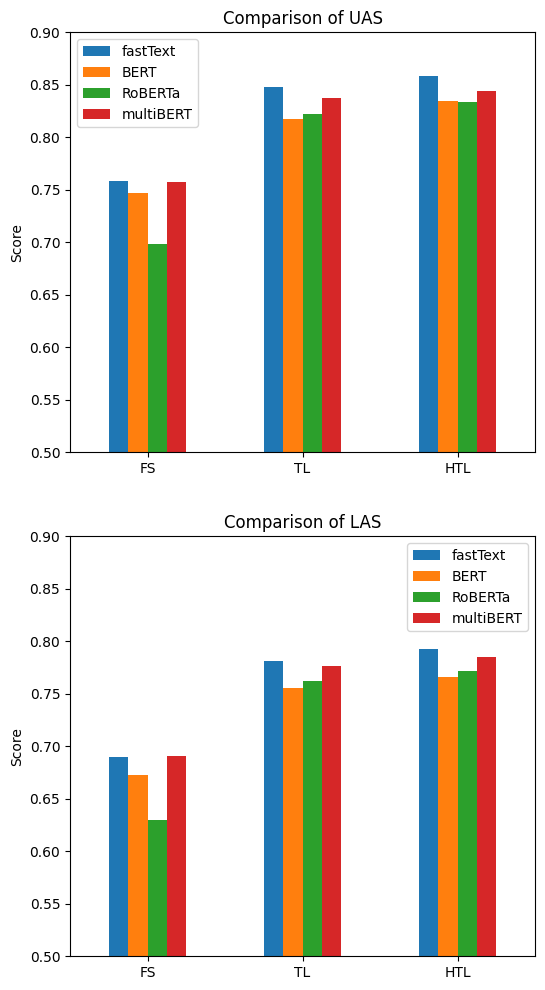}
	\caption{Comparison of the best model evaluation for each word embedding}
	\label{fig:compEmb}
\end{figure}

\subsection{Word Embeddings Comparison}

Figure \ref{fig:compEmb} shows that the model with a higher UAS score was obtained from word embedding fastText, followed by multilingual BERT, Javanese BERT, and Javanese RoBERTa. For LAS evaluation, the sequence is fastText, multilingual BERT, Javanese RoBERTa, and Javanese BERT. Although fastText is slightly superior, the differences are insignificant when considering the models' margin of error.

\subsection{Error Analysis}

\begin{table}[t!]
  \caption{Top 10 errors of the from-scratch model and its comparison with the transfer-learning model}
  \label{tab:error}
  \begin{tabular}{ccrrr}
    \toprule
    Ground Truth & Prediction & FS & TL & HTL\\
    \midrule

    obl & obj & 17 & 16 & 15 \\ 
    obl & nsubj & 7 & 3 & 7 \\ 
    obj & obl & 7 & 13 & 12 \\ 
    advcl & xcomp & 5 & 5 & 6 \\ 
    nmod & flat & 4 & 2 & 1 \\ 
    xcomp & advcl & 4 & 5 & 5 \\ 
    xcomp & obl & 3 & 3 & 2 \\ 
    nmod & obl & 3 & 1 & 1 \\ 
    nsubj & obj & 3 & 1 & 1 \\ 
    obj & nsubj & 2 & 0 & 3 \\

  \bottomrule
\end{tabular}
\end{table}

Table \ref{tab:error} displays more detail about the performance difference. The ten labels taken are obtained from pairs with the highest errors in the from-scratch model. Some pairs significantly reduce error, but there are also pairs with no significant changes and even more errors in scenarios with transfer learning.

One noteworthy insight is the significantly increasing error of words with "obj" label that predicted with "obl". It seems contradictory that model accuracy is increasing simultaneously with the addition of transfer learning. It turns out that there are a few differences in the word labeling of both labels between the source and the target language, so the model could not predict the word label correctly.

\section{Conclusions and Future Work}
\label{s:conclude}

This section explains the conclusion and improvements that can be developed from this work.

\subsection{Conclusions}

This work investigates whether cross-lingual transfer learning works for dependency parsing tasks of a low-resource language, Javanese. The result shows that the cross-lingual transfer learning model is significantly better than the baseline model. Models with transfer learning can improve performance on UAS and LAS metrics by up to 10\%. 

The best model was obtained from the hierarchical transfer learning method using Italian and English as the source and Indonesian as the intermediary languages. Meanwhile, the standard transfer learning method achieved the best accuracy using Indonesian as the source language. However, the differences between standard transfer learning and hierarchical learning are insignificant, considering the margin of error from each scenario.

\subsection{Future Work}

We focused more on the model's learning scheme than the model's development with the highest score. We use architecture from \citet{Dozat2017} rather than the one built by \citet{Mrini2020}, the state-of-the-art dependency parsing task. So, better architecture can be used to produce a model with a higher evaluation score in the future.

Our future works also include further error analysis, especially related to the languages involved that LangRank chose. It could investigate languages with different demography and characteristics (Croatian and Korean) compared to Javanese.

\section*{Limitations}

The following are the limitations of this research:

\begin{enumerate}

    \item There is no hyper-parameter tuning treatment in the model creation process.
    
    \item Cross-validation is not performed in the data distribution process. 

    \item Only one language is used as an intermediary language in hierarchical transfer learning.
    
\end{enumerate}

\section*{Acknowledgements}

We thank the Directorate of Research and Development, Universitas Indonesia, under Hibah PUTI 2022 (Grant No. NKB-1384/UN2.RST/HKP.05.00/2022) for funding this research. We also thank Tokopedia-UI AI Center, Faculty of Computer Science Universitas Indonesia, and Program Kompetisi Kampus Merdeka from the Ministry of Education, Culture, Research, and Technology Republic of Indonesia for the NVIDIA-DGX server we used for experiments. We also gratefully acknowledge Dr. Fajri Koto for the insightful feedback during the pre-submission mentorship.

\bibliography{custom}
\bibliographystyle{acl_natbib}

\appendix

% \section{Example Appendix}
% \label{sec:appendix}

% This is an appendix.

\end{document}